\documentclass[11pt]{article}
\usepackage{arxiv}
\usepackage[utf8]{inputenc}
\usepackage[T1]{fontenc}
\usepackage{hyperref}
\usepackage{url}
\usepackage{booktabs}
\usepackage{amsfonts}
\usepackage{nicefrac}
\usepackage{microtype}
\usepackage{graphicx}
\usepackage{listings}
\usepackage{xcolor}
\usepackage{amsmath}
\usepackage{float}

\lstdefinestyle{python}{
    language=Python,
    basicstyle=\ttfamily\small,
    keywordstyle=\color{blue},
    commentstyle=\color{gray},
    stringstyle=\color{green!60!black},
    breaklines=true,
    frame=single,
    backgroundcolor=\color{gray!8},
}

\title{FairHealth: An Open-Source Python Library for\\
Trustworthy Healthcare AI in Low-Resource Settings}

\author{
  Farjana Yesmin \\
  Independent Researcher \\
  Boston, MA, USA \\
  \texttt{farjanayesmin76@gmail.com} \\
}

\date{}

\begin{document}

\maketitle

\begin{abstract}
We present \textbf{FairHealth}, an open-source Python library that provides
a unified, modular framework for trustworthy machine learning in healthcare
applications, with particular focus on low-resource and low-income country
(LMIC) settings such as Bangladesh. FairHealth addresses four critical gaps
in existing healthcare AI toolkits: (1) the absence of integrated fairness
auditing for biosignals and clinical tabular data; (2) the lack of
privacy-preserving federated learning tools compatible with standard ML
workflows; (3) missing explainability tools tailored for low-bandwidth
clinical decision support; and (4) no existing toolkit covering
Global South healthcare datasets. Built from five peer-reviewed research
contributions, FairHealth provides six modules covering federated learning
with homomorphic encryption (\texttt{fairhealth.federated}), intersectional
fairness metrics (\texttt{fairhealth.fairness}), hybrid fuzzy-SHAP
explainability (\texttt{fairhealth.explain}), multilingual dengue triage
(\texttt{fairhealth.lowresource}), equitable disaster aid allocation
(\texttt{fairhealth.equity}), and public dataset loaders
(\texttt{fairhealth.datasets}). All datasets used are publicly available
without institutional data use agreements. FairHealth is installable via
\texttt{pip install fairhealth} and available at
\url{https://github.com/Farjana-Yesmin/fairhealth}.
\end{abstract}

\section{Introduction}

Machine learning has demonstrated substantial promise in healthcare
applications \cite{obermeyer2019}, yet three structural problems limit
its real-world impact, particularly in low-resource settings:

\textbf{Demographic bias.} Models trained on population-level data
frequently underperform for minority demographic groups. For ECG-based
myocardial infarction detection, uncorrected models achieve a disparate
impact ratio of 0.23 across sex groups --- well below the 0.80 threshold
considered equitable in algorithmic fairness literature \cite{feldman2015}.

\textbf{Privacy.} Healthcare records are legally protected in most
jurisdictions. Training ML models across hospitals without sharing raw
patient data requires federated learning, yet no existing Python library
provides federated learning with cryptographic homomorphic encryption (HE)
in a form accessible to healthcare ML researchers.

\textbf{Explainability in low-resource settings.} Clinical decision support
tools deployed in Bangladesh and similar LMIC settings must operate with
minimal connectivity, support local languages, and provide explanations
clinicians can interpret without ML expertise. Existing explainability
libraries (SHAP, LIME) provide no clinical workflow integration.

Existing healthcare AI toolkits such as PyHealth \cite{pyhealth2023} provide
broad coverage of EHR datasets and clinical tasks but do not address
fairness auditing, federated learning, or LMIC-specific deployment. FairHealth
is designed as a complementary layer: it focuses exclusively on the
trustworthiness dimension that PyHealth deliberately leaves open.

FairHealth makes the following contributions:

\begin{enumerate}
\item A unified, pip-installable Python library with six modules spanning
      federated learning, fairness, explainability, low-resource tools,
      equity, and datasets.
\item The first healthcare AI toolkit built entirely on publicly available
      datasets, requiring no institutional data use agreements.
\item A curated collection of Bangladesh-specific health datasets
      (maternal health, dengue surveillance, flood PDNA) not available
      in any existing ML library.
\item Open implementations of five peer-reviewed methods, enabling
      reproducibility and extension.
\end{enumerate}

\section{Related Work}

\textbf{Healthcare AI toolkits.}
PyHealth \cite{pyhealth2023} is the most comprehensive open-source healthcare
ML library, covering 20+ EHR datasets and 33+ clinical models. However,
it does not include fairness metrics, federated learning, or differential
privacy. FATE \cite{fate2021} provides federated learning infrastructure
but is not healthcare-specific and requires significant engineering overhead.
IBM AIF360 \cite{aif360_2018} provides fairness metrics but does not
integrate with healthcare-specific datasets or federated workflows.
FairHealth fills the intersection of these three spaces.

\textbf{Trustworthy AI for LMICs.}
Healthcare AI research overwhelmingly focuses on datasets from North America
and Europe --- MIMIC-III, eICU, UK Biobank --- which require institutional
data use agreements inaccessible to independent researchers. FairHealth is
the first toolkit to curate and standardize openly accessible health datasets
from South Asia, including Bangladesh maternal health records, dengue
surveillance data, and official government flood damage assessments.

\section{Library Design}

\subsection{Architecture}

FairHealth follows a modular architecture where each submodule corresponds
to a distinct research contribution (Figure 1). Modules are loosely coupled:
a user can import only \texttt{fairhealth.fairness} without installing the
federated learning dependencies.

\begin{lstlisting}[style=python]
pip install fairhealth           # core
pip install "fairhealth[federated]"  # adds TenSEAL for HE
pip install "fairhealth[explain]"    # adds SHAP, LIME, skfuzzy
pip install "fairhealth[all]"        # everything
\end{lstlisting}

\subsection{Design Principles}

\textbf{Public data only.} Every dataset loader in
\texttt{fairhealth.datasets} downloads from publicly available sources.
No institutional affiliation or DUA is required. This is a deliberate design
choice enabling reproducibility for independent researchers in any country.

\textbf{Paper-anchored modules.} Each module is anchored to a specific
peer-reviewed publication, with the paper's key results documented in the
module docstring. This enables users to trace every implementation decision
to a citable source.

\textbf{Clinical framing.} Fairness metrics, explanations, and triage
outputs are framed in clinical language rather than ML jargon, following
feedback from the 14-clinician validation study documented in
\cite{yesmin2026icaihe}.

\section{Modules}

\subsection{\texttt{fairhealth.fairness} — Fairness Metrics for Biosignals}

\textbf{Motivation.} ECG-based disease prediction models in wearable systems
exhibit significant demographic bias. Evaluated on the PTB-XL dataset
\cite{ptbxl2020} (4,367 records, 20\% subsample), an uncorrected CNN
classifier achieves disparate impact (DI) of 0.23 across sex groups ---
far below the equitable threshold of 0.80. After adversarial debiasing
using a gradient reversal layer \cite{ganin2016}, DI improves to 0.71
while AUROC is maintained at 0.8472 \cite{yesmin2026mobih}.

\textbf{Implementation.} The module provides:

\begin{lstlisting}[style=python]
from fairhealth.fairness.metrics import (
    demographic_parity_diff,   # DPD
    equalized_odds_diff,       # EOD
    disparate_impact,          # DIR
    intersectional_fairness,   # multi-attribute
    fairness_summary,          # all metrics in one call
)

# Example: audit ECG model across sex groups
dpd = demographic_parity_diff(y_pred, sensitive=sex_array)
# Returns float in [0,1]; 0 = perfect parity
\end{lstlisting}

All metrics accept numpy arrays and are model-agnostic. The
\texttt{intersectional\_fairness} function extends standard parity
metrics to multiple simultaneous sensitive attributes (e.g.,
sex $\times$ age group), addressing the intersectionality gap identified
in \cite{yesmin2026mobih}.

\subsection{\texttt{fairhealth.explain} — Hybrid Fuzzy-XGBoost Explainability}

\textbf{Motivation.} Black-box models create trust deficits in clinical
settings, particularly in resource-constrained environments where clinicians
cannot consult ML specialists \cite{rudin2019}. A clinician validation study
(N=14 healthcare professionals) demonstrated that 71.4\% preferred the
hybrid Fuzzy+SHAP explanation over SHAP-only (24\%) or score-only (5\%)
explanations across three clinical cases \cite{yesmin2026icaihe}.

\textbf{Implementation.} The hybrid Fuzzy-XGBoost model achieves 88.67\%
accuracy (ROC-AUC=0.9703) on the UCI Maternal Health Risk dataset
\cite{uci_maternal}, outperforming the best baseline (Gradient Boosting:
86.21\%) by 2.46 percentage points. The module provides both ante-hoc
(fuzzy rules) and post-hoc (SHAP) explanations:

\begin{lstlisting}[style=python]
from fairhealth.explain.fuzzy import get_fired_rules, score_to_label

# Ante-hoc explanation: which clinical rules fired?
rules = get_fired_rules(age=42, sbp=145, bs=12.0, hr=88)
for r in rules:
    print(f"Rule {r['id']}: {r['condition']} -> {r['outcome']}")
# Rule 1: High BP AND High Blood Sugar -> HIGH RISK
# Rule 5: High Heart Rate AND High BP  -> HIGH RISK
\end{lstlisting}

Fairness analysis revealed equitable regional performance
($\sigma$=0.0766 across 8 Bangladesh divisions), with a counter-intuitive
negative correlation (r=$-$0.876) between healthcare access score and
model accuracy --- suggesting the model performs best precisely where
specialist expertise is most scarce \cite{yesmin2026icaihe}.

\subsection{\texttt{fairhealth.federated} — Privacy-Preserving Federated Learning}

\textbf{Motivation.} Sharing patient data across hospitals is legally
prohibited in most jurisdictions. Standard federated learning (FedAvg
\cite{mcmahan2017}) transmits gradient updates that remain vulnerable to
membership inference attacks (MIA), with a worst-case attack success rate
of 56.3\% in standard FL \cite{yesmin2026medhe}.

\textbf{Implementation.} MedHE co-designs adaptive gradient sparsification
with CKKS homomorphic encryption \cite{cheon2017}. Transmitting only the
top 10\% of gradient magnitudes packed into CKKS ciphertexts reduces
communication from 1,277 MB to 32 MB (97.5\% reduction) while maintaining
macro-F1=0.950$\pm$0.005, statistically equivalent to standard FedAvg
(p=0.32). MIA resistance improves to 51.1\% (near-random, ideal=50\%)
\cite{yesmin2026medhe}:

\begin{lstlisting}[style=python]
from fairhealth.federated.privacy import (
    clip_weights,           # L2 norm clipping
    add_gaussian_noise,     # Gaussian DP (epsilon, delta)
    sparsify,               # adaptive gradient sparsification
    dp_fedavg_aggregate,    # full DP-FedAvg pipeline
)

# 97.5% communication reduction
sparse_w, rate = sparsify(weights, sparsity=0.975)
# Strong differential privacy: epsilon=1.0
noisy_w = add_gaussian_noise(clipped_w, epsilon=1.0)
\end{lstlisting}

\subsection{\texttt{fairhealth.lowresource} — Multilingual Dengue Triage}

\textbf{Motivation.} Bangladesh reported 321,179 dengue cases and 1,705
deaths in 2023 --- the deadliest outbreak since 2000 \cite{araf2024}.
Healthcare facilities become overwhelmed during outbreaks, creating demand
for AI-powered preliminary triage that operates in low-bandwidth conditions
and supports the Bengali language.

\textbf{Implementation.} The module implements a Decision Tree classifier
trained on demographic features (Age, Gender, AreaType, HouseType, District),
achieving Accuracy=0.79, F1=0.802, AUC=0.851 on non-leaky features. Age is
the dominant predictor (Gini importance=0.686), with District and HouseType
as secondary signals confirmed by SHAP analysis \cite{yesmin2026dasgri}.
The confidence threshold mechanism (P$<$0.70 $\rightarrow$ reroute to
doctor) achieved 75\% user satisfaction in a pilot study (n=50):

\begin{lstlisting}[style=python]
from fairhealth.lowresource.triage import assess_dengue_risk

# English or Bangla output
result = assess_dengue_risk(
    age=8, gender="male", area_type="urban",
    district="Dhaka", language="bangla"
)
# {"prediction": "Severe",
#  "recommendation": "Seek immediate medical attention"}
\end{lstlisting}

\subsection{\texttt{fairhealth.equity} — Equitable Disaster Aid Allocation}

\textbf{Motivation.} Post-disaster aid allocation in Bangladesh
systematically underserves rural Haor regions despite their higher flood
vulnerability. The 2022 Bangladesh floods affected 7.2 million people and
caused \$405.5M in damages across 11 districts \cite{pdna2023}, yet
standard AI models trained on historical allocation data perpetuate
existing urban biases.

\textbf{Implementation.} The adversarial debiasing architecture employs a
gradient reversal layer to learn district-invariant vulnerability
representations. Evaluated on 87 upazilas from the official PDNA dataset,
the fair model reduces statistical parity difference by 41.6\% and regional
fairness gap by 43.2\%, with only a 2.7 percentage point R$^2$ cost
(0.784 vs 0.811 baseline) \cite{yesmin2026ccai}. Priority rankings
shift substantially: 70.6\% of upazilas receive different rankings,
with Sunamganj (42.7\% poverty rate, \$159.6M damage) moving from
rank 14 to rank 6:

\begin{lstlisting}[style=python]
from fairhealth.equity.flood_aid import generate_priority_ranking

rankings = generate_priority_ranking(verbose=True)
# Rank 1: Sunamganj (priority=0.9428, Haor region)
# Rank 2: Sylhet    (priority=0.6062, Haor region)
\end{lstlisting}

\subsection{\texttt{fairhealth.datasets} — Public Dataset Loaders}

All dataset loaders download data at runtime to a local cache
(\texttt{\textasciitilde/.fairhealth/data/}). No institutional affiliation,
hospital DUA, or special credentials are required for any dataset in
Table~\ref{tab:datasets}.

\begin{table}[H]
\centering
\caption{Datasets available in \texttt{fairhealth.datasets}}
\label{tab:datasets}
\begin{tabular}{llll}
\toprule
\textbf{Dataset} & \textbf{Domain} & \textbf{Size} & \textbf{Source} \\
\midrule
PTB-XL & ECG biosignals & 21,837 records & PhysioNet (free) \\
UCI Drug Reviews & NLP & 215,063 reviews & UCI ML Repo \\
Maternal Health Risk & Maternal risk & 1,014 records & UCI / Kaggle \\
Bangladesh Dengue & Surveillance & 4,700 records & Kaggle + DGHS \\
Bangladesh PDNA 2022 & Flood equity & 87 upazilas & Gov. open data \\
MIMIC-III Demo & EHR & 100 patients & PhysioNet (free acct) \\
\bottomrule
\end{tabular}
\end{table}

\section{Comparison With Related Libraries}

Table~\ref{tab:comparison} positions FairHealth relative to existing
healthcare AI and fairness toolkits.

\begin{table}[H]
\centering
\caption{Feature comparison with related libraries}
\label{tab:comparison}
\begin{tabular}{lccccc}
\toprule
\textbf{Feature} & \textbf{FairHealth} & \textbf{PyHealth} &
\textbf{AIF360} & \textbf{FATE} \\
\midrule
Fairness metrics       & \checkmark & $\times$ & \checkmark & $\times$ \\
Federated learning     & \checkmark & $\times$ & $\times$   & \checkmark \\
Homomorphic encryption & \checkmark & $\times$ & $\times$   & $\times$ \\
Differential privacy   & \checkmark & $\times$ & $\times$   & \checkmark \\
Fuzzy-SHAP explainability & \checkmark & $\times$ & $\times$ & $\times$ \\
Low-resource / LMIC focus & \checkmark & $\times$ & $\times$ & $\times$ \\
Bangladesh datasets    & \checkmark & $\times$ & $\times$   & $\times$ \\
Multilingual support   & \checkmark & $\times$ & $\times$   & $\times$ \\
No DUA required        & \checkmark & $\times$ & \checkmark & \checkmark \\
\bottomrule
\end{tabular}
\end{table}

\section{Installation and Usage}

FairHealth requires Python 3.9+ and is tested on Python 3.9--3.12.

\begin{lstlisting}[style=python]
pip install fairhealth

import fairhealth as fh
import numpy as np

# Complete trustworthy AI pipeline in 10 lines
from fairhealth.fairness.metrics import demographic_parity_diff
from fairhealth.explain.fuzzy    import get_fired_rules
from fairhealth.lowresource.triage import assess_dengue_risk
from fairhealth.equity.flood_aid   import generate_priority_ranking
from fairhealth.federated.privacy  import sparsify

# 1. Fairness audit
dpd = demographic_parity_diff(y_pred, sensitive)

# 2. Clinical explanation
rules = get_fired_rules(age=42, sbp=145, bs=12.0, hr=88)

# 3. Dengue triage (Bangla)
result = assess_dengue_risk(8, "male", "urban", "Dhaka",
                             language="bangla")

# 4. Equitable aid ranking
rankings = generate_priority_ranking(verbose=False)

# 5. Communication-efficient federated learning
sparse_w, rate = sparsify(weights, sparsity=0.975)
\end{lstlisting}

\section{Conclusion}

FairHealth provides the first unified Python library for trustworthy
healthcare AI that simultaneously addresses fairness, privacy, and
explainability, with a specific focus on low-resource and LMIC settings.
Its six modules are each anchored to peer-reviewed research, ensuring
every implementation is traceable, reproducible, and citable. By relying
exclusively on publicly available datasets, FairHealth enables researchers
worldwide --- including those without institutional hospital access ---
to conduct rigorous healthcare AI research.

Future work will expand the federated module to include full TenSEAL-based
CKKS encryption for neural network weight matrices, add the PTB-XL
adversarial debiasing model as a trained artifact, and extend the dengue
module with real-time DGHS dashboard integration.

FairHealth is available at \url{https://github.com/Farjana-Yesmin/fairhealth}
and installable via \texttt{pip install fairhealth}.

\section*{Acknowledgements}
The author thanks the 14 healthcare professionals who participated in
the clinician validation survey, the Government of Bangladesh for making
PDNA and DGHS data publicly available, and the maintainers of the
UCI ML Repository, PhysioNet, and Kaggle for hosting open health datasets.

\bibliographystyle{unsrt}
\bibliography{references}

\end{document}